\documentclass{article}

\usepackage{PRIMEarxiv}

\usepackage[utf8]{inputenc} 
\usepackage[T1]{fontenc}    
\usepackage{hyperref}       
\usepackage{url}            
\usepackage{booktabs}       
\usepackage{amsfonts}       
\usepackage{nicefrac}       
\usepackage{microtype}      
\usepackage{lipsum}
\usepackage{multirow}
\usepackage{fancyhdr}       
\usepackage{graphicx}       
\graphicspath{{media/}}     

\title{Enhancing Construction Site Safety: A Lightweight Convolutional Network for Effective Helmet Detection}

\author{
  Mujadded Al Rabbani Alif \\
  Department of Computer Science \\
  Huddersfield University \\
  Queensgate, Huddersfield HD1 3DH, UK\\
  m.alif@hud.ac.uk \\
}

\begin{document}
\maketitle

\begin{abstract}
In the realm of construction safety, the detection of personal protective equipment, such as helmets, plays a critical role in preventing workplace injuries. This paper details the development and evaluation of convolutional neural networks (CNNs) designed for the accurate classification of helmet presence on construction sites. Initially, a simple CNN model comprising one convolutional block and one fully connected layer was developed, yielding modest results. To enhance its performance, the model was progressively refined, first by extending the architecture to include an additional convolutional block and a fully connected layer. Subsequently, batch normalization and dropout techniques were integrated, aiming to mitigate overfitting and improve the model's generalization capabilities. The performance of these models is methodically analyzed, revealing a peak F1-score of 84\%, precision of 82\%, and recall of 86\% with the most advanced configuration of the first study phase. Despite these improvements, the accuracy remained suboptimal, thus setting the stage for further architectural and operational enhancements. This work lays a foundational framework for ongoing adjustments and optimization in automated helmet detection technology, with future enhancements expected to address the limitations identified during these initial experiments.
\end{abstract}


\section{Introduction}
\begin{figure}
  \centering
  \includegraphics[width=0.7\columnwidth]{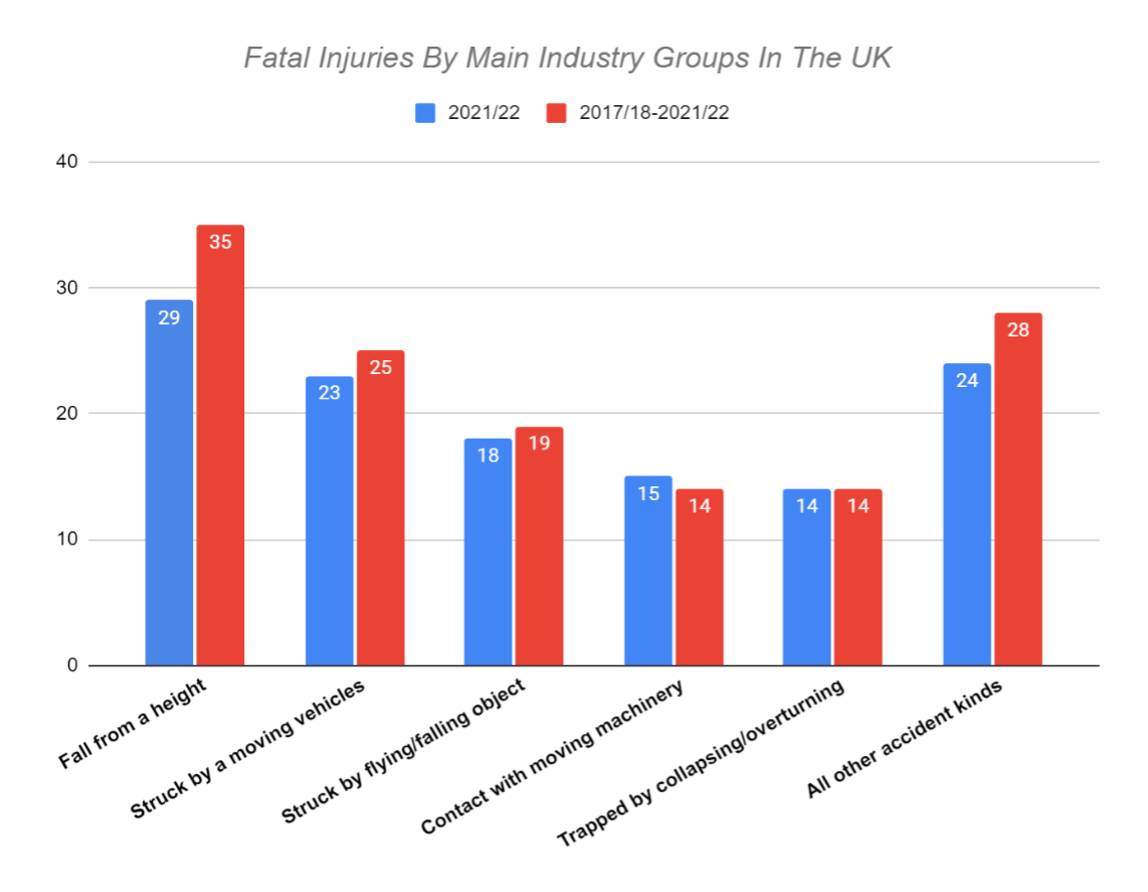}
  \caption{Comparison of annual fatal injuries across major industry sectors for 2021/22 with averages from the five-year period 2017/18 to 2021/22 \cite{RN3}}
  \label{fig:fig1}
\end{figure}

The construction industry is notoriously fraught with hazards, characterized by high-risk activities and environments that substantially increase the likelihood of worker exposure to dangerous situations. Recent statistical data reveals a distressing global trend of rising fatalities within this sector. Specifically, the U.S. Bureau of Labor Statistics reported that the year 2021 saw a total of 5,190 fatal work injuries, which represents an 8.9\% increase from the preceding year \cite{RN1}. This upward trajectory is mirrored in China, where the Ministry of Emergency Management noted a 7.8\% increase in construction-related accidents and a 1.4\% rise in fatalities during the first half of 2018. Contrary to the decreasing accident and fatality rates observed in most other industries, the construction sector has experienced a consistent increase in such incidents since 2016 \cite{RN2}. Furthermore, the UK Health and Safety Executive (HSE) highlights a particularly alarming statistic: 79\% of all fatal injuries in the construction industry from 2017/18 to 2021/22 were confined to just five types of accidents, as illustrated in Figure-\ref{fig:fig1}. Within the most recent year of the study (2021/22), falls from height alone were responsible for 29 fatalities, constituting 24\% of all construction worker deaths for that year \cite{RN3}. These figures underscore a critical need for targeted interventions to address specific high-risk activities within the construction environment. The persistent elevation in fatality rates, despite technological and regulatory advancements, calls for innovative approaches to enhance safety measures and compliance. This paper explores the potential of leveraging advanced machine learning techniques and computer vision to improve safety equipment compliance and reduce the occurrence of fatal accidents in the construction industry.

Proper utilization of personal protective equipment (PPE), such as helmets, gloves, and safety glasses, plays a pivotal role in mitigating injuries and fatalities within the workplace \cite{RN4}. Helmets are particularly essential, safeguarding construction workers from head injuries resulting from impacts with falling objects or direct collisions. Research indicates that helmets can reduce the risk of severe brain injury by up to 95\% when struck by objects like concrete blocks \cite{RN5}. Nonetheless, reliance on manual inspection methods to ensure helmet usage and compliance with safety standards presents significant challenges. These traditional techniques are not only time-intensive but also prone to human errors, making them less effective on large construction sites where monitoring all workers consistently proves to be a daunting task \cite{RN6}. This inefficiency underscores the necessity for more reliable and automated systems to enhance compliance and safety oversight.

This paper addresses the critical issue of worker safety on construction sites by proposing an innovative automated helmet detection system. Utilizing Convolutional Neural Network (CNN) \cite{schmidhuber2015deep} technology for image classification, the system analyzes images from cameras installed around the site to determine whether workers are wearing helmets. The versatility of CNN architectures extends beyond helmet detection, effectively addressing challenges in various domains. For example, CNNs have been successfully applied to detect damages in pallet racking systems \cite{alif2024attention}, recognize isolated handwritten characters \cite{alif2017isolated, alif2024state}, and identify missing bolts in railway components \cite{alif2024lightweight, alif2024boltvision}. Moreover, these architectures have made significant inroads into healthcare, enhancing diagnostic procedures for conditions like diabetic retinopathy \cite{alif2024enhancing}. The integration of YOLO models further expands the potential of CNNs, especially in real-time applications, as demonstrated in agricultural monitoring \cite{alif2024yolov1} and vehicle detection \cite{sundaresan2024comparative}. The proposed CNN model employed here is meticulously trained on a substantial and diverse dataset, enabling it to accurately recognize the presence or absence of helmets under various conditions, including challenging lighting scenarios and different helmet types.

Significant contributions of this paper include the creation of a comprehensive, annotated dataset that reflects a wide range of real-world scenarios where workers may or may not wear helmets. Additionally, we introduce a specialized CNN architecture comprising three convolutional layers and three fully connected layers, incorporating max pooling to enhance feature aggregation and reduce positional dependencies. The model benefits from advanced training strategies such as data augmentation, rigorous regularization, and meticulous hyperparameter tuning.

A detailed evaluation of the system's performance is conducted, presenting key metrics such as accuracy, precision, recall, F1-score, and a confusion matrix. The findings and methodologies put forth in this paper significantly advance the field of PPE detection through CNN-based image classification, offering tailored domain-specific augmentations, an effective architecture, and a comprehensive training and validation approach that underscores the potential of this technology in enhancing safety compliance in the construction industry.

\section{Literature Review}
Machine learning-based object detection technologies have become increasingly prevalent across various domains due to their robust capabilities in detecting and classifying objects \cite{RN8}. In a seminal work by Rubaiyat et al. \cite{RN7}, an automatic detection method was introduced that effectively identifies both construction workers and safety helmets. This method utilizes a combination of frequency domain analysis, Histogram of Oriented Gradients (HOG), and Circle Hough Transform (CHT) in a sequential approach to enhance the accuracy of detecting workers equipped with helmets. Despite the successes, these machine learning-driven approaches encounter challenges, particularly in distinguishing safety helmets from other headgear, such as hats with similar colors and shapes. Moreover, the detection accuracy falters in scenarios where the workers’ faces are partially obscured, complicating the simultaneous recognition of faces and helmets.

Historically, research in this field has been oriented toward binary classification models that simply ascertain the presence or absence of safety helmets. Such models often overlook the presence of other forms of headwear like caps or headscarves, leading to potential misclassifications in diverse working environments \cite{RN9}. Addressing these limitations, Cheng et al. \cite{RN10} developed a more nuanced model that classifies headgear into four categories: helmet, cap, no-wear, and safety cap. This model, an enhanced version of YOLOv3-Tiny named SAS-YOLOv3-Tiny, integrates sandglass-residual structures and Spatial Pyramid Pooling (SPP) modules to strike a balance between detection accuracy and processing speed. Trained on a dataset comprising 7,656 images, their model demonstrated commendable performance metrics, achieving precision, recall, mean Average Precision (mAP), and F1-score of 71.6\%, 80.9\%, 80.3\%, and 75.2\% respectively.

Building upon the development of object detection models, a recent study by Z. Xiang et al. \cite{RN11} evaluated the effectiveness of various configurations of the YOLOv5 model for safety helmet detection. Modifications to the YOLOv5 models included adjustments to the size of the BottleneckCSP module, which is integral to the model's architecture. Among the variants, YOLOv5s demonstrated superior processing speed, achieving 110 frames per second despite similar mean Average Precision (mAP) values among the models. The addition of pre-training weights further enhanced the YOLOv5 models, improving mAP values by approximately 0.9 to 1.3 points. Another advancement in the YOLOv5 architecture was made by J. Doe et al. \cite{RN12}, who incorporated a multi-scale detection approach and the DIoU-NMS technique to refine the accuracy of bounding box predictions, especially for smaller targets. This modified YOLOv5 model recorded an impressive mAP of 95.7\% while maintaining a processing speed of 98 FPS.

The use of pre-trained convolutional neural networks (CNNs) such as VGG \cite{RN13}, Inception\_V3 \cite{RN14}, and ResNet50 \cite{RN15} has been prevalent in general image classification tasks, owing to their robust performance metrics established through training on large datasets like ImageNet \cite{RN16}. Inspired by these successes, K. Zdenek et al. \cite{RN17} adapted the VGG-16 model for a specialized task of guardrail detection. By retraining VGG-16 on a dataset of 4,000 augmented images, they effectively transferred learned image feature knowledge to a new domain. The features extracted were subsequently processed using an MLP model, culminating in a detection accuracy of 96.5\%, which notably exceeded the performance of a traditional support vector machine (SVM) approach.

Moreover, addressing the challenges posed by low-resolution objects in diverse scenarios, another pivotal study \cite{RN18} utilized the Faster R-CNN framework \cite{RN19}, employing the VGG-16 architecture as the backbone for its classification network. This approach proved highly effective for monitoring hard hat usage in far-field surveillance videos, demonstrating consistently high precision and recall rates above 90\% across varying weather conditions and worker poses.

In summary, convolutional neural networks (CNNs) have become the predominant technology for helmet detection in image classification tasks, outperforming traditional detection methods. Research has extensively demonstrated the efficacy of CNN architectures such as VGG-16 and Faster R-CNN in identifying safety helmets within complex construction environments. These networks excel due to their robust ability to process low-resolution images and adapt effectively to diverse conditions, making them ideally suited for the challenging scenarios often encountered in safety compliance monitoring. Furthermore, these methodologies have consistently delivered high precision and recall rates, maintaining robust performance amidst varying environmental factors and worker movements. Consequently, the application of CNNs in helmet detection not only enhances the reliability and accuracy of results but also underscores their superiority in managing complex and variable visual data, thereby affirming their vital role in advancing object detection and classification technologies in safety-critical applications.
\section{Methodology}
\subsection{Dataset}
A comprehensive dataset forms the foundation of our research on automated helmet detection. To ensure a robust and representative sample, we compiled images from two distinct sources: the publicly accessible "Safety Helmet Detection" dataset available on Kaggle \cite{RN20} and direct image acquisitions from various operational construction sites. This dual-source approach allows for a diverse collection of images that encompasses both staged digital representations and dynamic real-life scenarios.

The ethical integrity of our research was maintained by adhering strictly to relevant data collection protocols and privacy regulations. Explicit consent was obtained from all individuals involved, including workers and, where applicable, their legal guardians. This consent process ensures the protection of participants' privacy and compliance with ethical standards.

The dataset consists of 500 high-quality images carefully selected to represent diverse conditions typical of construction sites. These images are categorized into two primary classes based on the presence or absence of safety helmets. The distribution of these classes is balanced to prevent any model bias during the training process. The following table \ref{tab:table1} provides a detailed breakdown of the dataset composition:

\begin{table}[h]
\centering
\caption{Breakdown of Image Classes in the Helmet Detection Dataset}
\begin{tabular}{|c|c|}
\hline
\textbf{Class} & \textbf{Samples} \\
\hline
Helmet & 250 \\
\hline
No Helmet & 250 \\
\hline
\end{tabular}
\label{tab:table1}
\end{table}

This balanced dataset facilitates the training of our convolutional neural network (CNN), enabling it to learn distinctive features associated with helmets and to generalize well across different operational environments encountered on construction sites.

\subsection{Dataset Visualization and Partitioning}
\begin{figure}[h]
    \centering
    {\includegraphics[width=0.5\columnwidth]{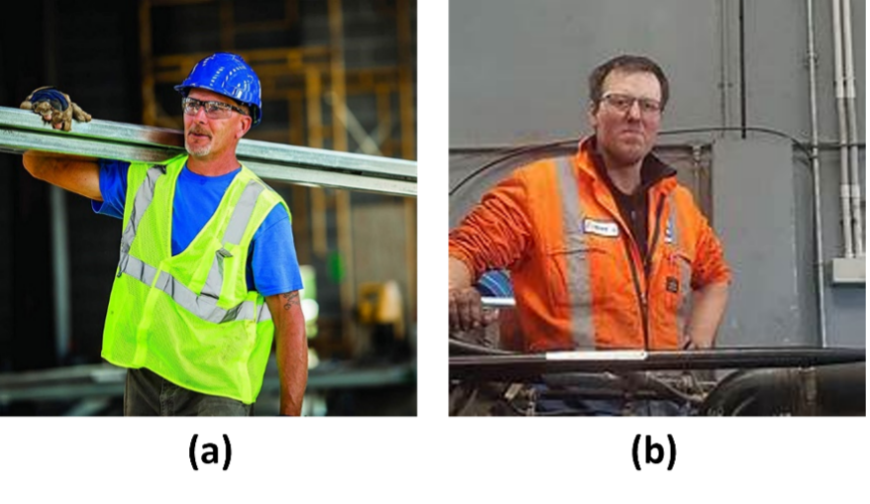}}
    \caption{Visual Comparison of Construction Site Workers: (a) Equipped with Safety Helmet, (b) Not Wearing Safety Helmet.}
    \label{fig:figure2}
\end{figure}

Figure \ref{fig:figure2} provides a visual representation of the dataset, showcasing examples of workers both with and without helmets. This visual comparison is integral for discerning the unique features that differentiate the two classes. Key attributes such as the presence, shape, color, and position of the helmet were meticulously analyzed to aid the development of an effective image classification model.

\begin{table}[!htbp]
\centering
\caption{Distribution of Helmet Detection Dataset by Usage}
\begin{tabular}{|c|c|c|c|}
\hline
\textbf{Subset} & \textbf{With Helmet} & \textbf{Without Helmet} & \textbf{Total Samples} \\
\hline
Training (68\%) & 170 & 180 & 350 \\
\hline
Validation (22\%) & 55 & 45 & 100 \\
\hline
Testing (10\%) & 28 & 22 & 50 \\
\hline
\end{tabular}
\label{tab:table2}
\end{table}

This analysis was conducted prior to the application of any augmentation techniques, establishing a clear baseline understanding of the dataset's characteristics. Although the dataset comprises only 500 images, it effectively represents the diverse scenarios typically encountered on construction sites. To systematically train and evaluate the performance of the CNN model, the dataset was partitioned into three subsets: training, validation, and testing. These subsets were distributed using a 70:20:10 ratio, respectively, facilitated by the `train-test-split` function. Detailed information on this partitioning is summarized in Table \ref{tab:table2}. This strategic division ensures a comprehensive training and evaluation process, enhancing the model's ability to generalize well across unseen data.

\subsection{Data Augmentation}
To mitigate the risks posed by overfitting and to enhance the model's generalization capabilities, several data augmentation techniques were applied to the original dataset. Each technique was specifically chosen to introduce realistic variability into the training process that mirrors actual conditions on construction sites.

\subsubsection{Crop Augmentation}
\begin{figure}[h]
    \centering
    \includegraphics[width=0.5\linewidth]{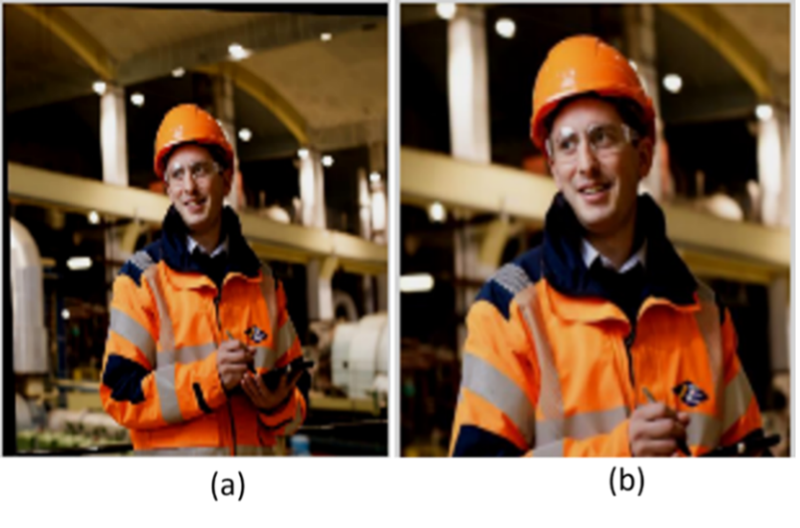}
    \caption{Illustration of Crop Augmentation Technique Showing 35\% Reduction: (a) Original Image, (b) Post-Augmentation}
    \label{fig:figure3}
\end{figure}

Crop augmentation was implemented to vary image composition and emphasize critical regions within the frame. Figure \ref{fig:figure3} showcases the application of a 35\% crop, which focuses on different positions and scales of the helmets. Such augmentation aids the model in recognizing helmets across varied perspectives and distances.

\subsubsection{Rotation Augmentation}
\begin{figure}[h]
    \centering
    \includegraphics[width=0.75\linewidth]{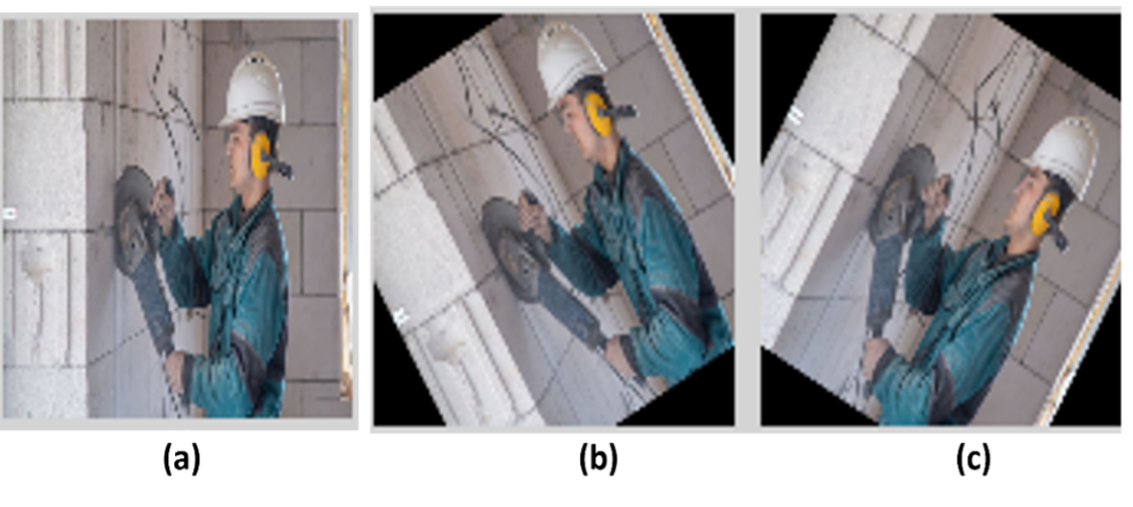}
    \caption{Demonstration of 30° Rotation Augmentation on an Image: (a) Original, (b) Rotated Counterclockwise by 30°, (c) Rotated Clockwise by 30°}
    \label{fig:figure4}
\end{figure}

\begin{figure}[h]
    \centering
    \includegraphics[width=0.75\linewidth]{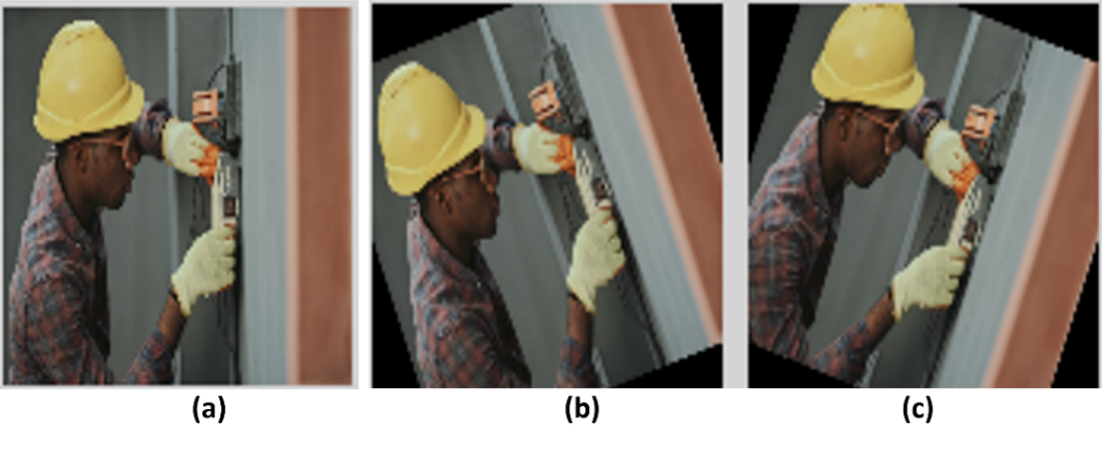}
    \caption{Example of 20° Rotation Augmentation: (a) Original Image, (b) Rotated Counterclockwise 20°, (c) Rotated Clockwise 20°}
    \label{fig:figure5}
\end{figure}

To address the challenges of skewed helmet positions due to worker movement or varied camera angles, rotation augmentation was utilized. Figures \ref{fig:figure4} and \ref{fig:figure5} depict rotations of 30° and 20°, respectively. This method trains the model to identify helmets in tilted or skewed states, enhancing its robustness.

\begin{figure}[!htbp]
    \centering
    \includegraphics[width=0.75\linewidth]{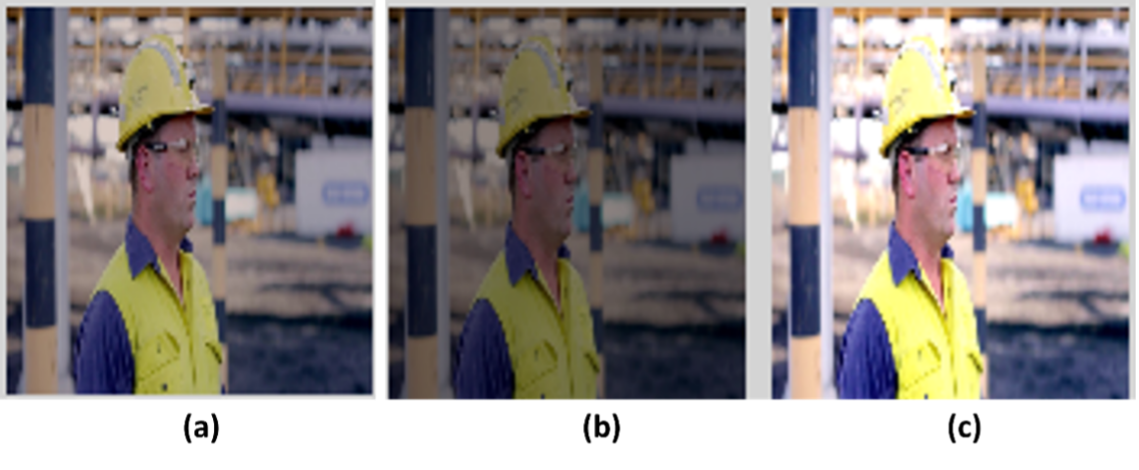}
    \caption{Adjustments in Image Brightness by 35\%: (a) Original Image, (b) Reduced Brightness by 35\%, (c) Increased Brightness by 35\%}
    \label{fig:figure6}
\end{figure}

\begin{figure}[h]
    \centering
    \includegraphics[width=0.75\linewidth]{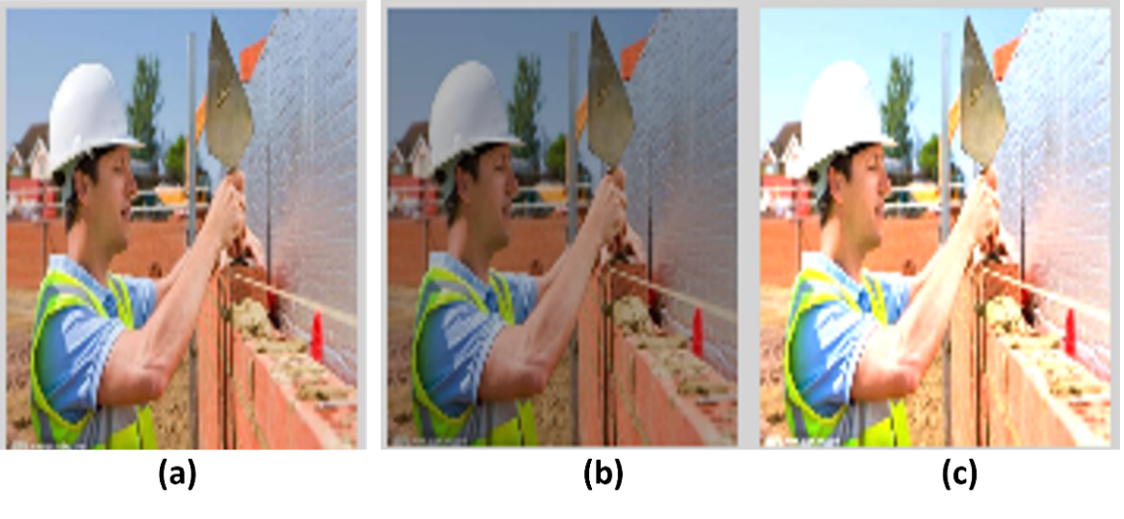}
    \caption{Implementation of 28\% Brightness Augmentation: (a) Original Image, (b) Decreased Brightness by 28\%, (c) Increased Brightness by 28\%}
    \label{fig:figure7}
\end{figure}

\subsubsection{Brightness Augmentation}
To ensure that the model performs well under various lighting conditions, brightness augmentation was employed. Figures \ref{fig:figure6} and \ref{fig:figure7} illustrate adjustments of 35\% and 28\% in brightness levels, respectively. This adjustment enables the model to maintain accuracy even under significant lighting variations.

Post augmentation, the dataset expanded to a total of 2886 images. The augmented data was then divided into training, testing, and validation subsets, maintaining the original distribution ratio to ensure a consistent and fair evaluation process. The detailed breakdown of the augmented dataset is presented in Table \ref{tab:table3}.

\begin{table}[h]
\centering
\caption{Distribution of Samples in the Augmented Helmet Detection Dataset}
\begin{tabular}{|c|c|c|c|}
\hline
\textbf{Subset} & \textbf{Helmet} & \textbf{No Helmet} & \textbf{Total Samples} \\
\hline
Training (70\%) & 975 & 1045 & 2020 \\
\hline
Validation (20\%) & 283 & 294 & 577 \\
\hline
Testing (10\%) & 134 & 155 & 289 \\
\hline
\end{tabular}
\label{tab:table3}
\end{table}

\subsection{Proposed Architecture}
Initially, our architecture comprised a single convolutional block and one fully connected layer, with the convolutional block outputting 11 channels and the fully connected layer having 40 neurons. This initial design did not perform satisfactorily on both the original and augmented datasets, prompting further iterations to refine the model.

After several modifications, the architecture evolved to include two convolutional blocks followed by two fully connected layers. The first convolutional block produces 11 output channels, and the second outputs 22. Each block uses ReLU activation and is followed by max-pooling to reduce spatial dimensions. The fully connected layers are designed with decreasing neuron counts, starting from 100, reducing to 50, and ultimately leading to a 2-neuron output layer for classification. Despite these enhancements, performance improvements were necessary.

\subsubsection{Final Model Architecture}
The final architecture that we propose is a lightweight model composed of three convolutional blocks followed by three fully connected layers optimized for efficient helmet detection. The layers are arranged as follows:
- The first convolutional block filters the input image into 11 output channels,
- The second block increases the complexity with 22 output channels,
- The third block doubles the capacity to 44 output channels.

\begin{figure}[!htbp]
    \centering
    \includegraphics[width=0.75\linewidth]{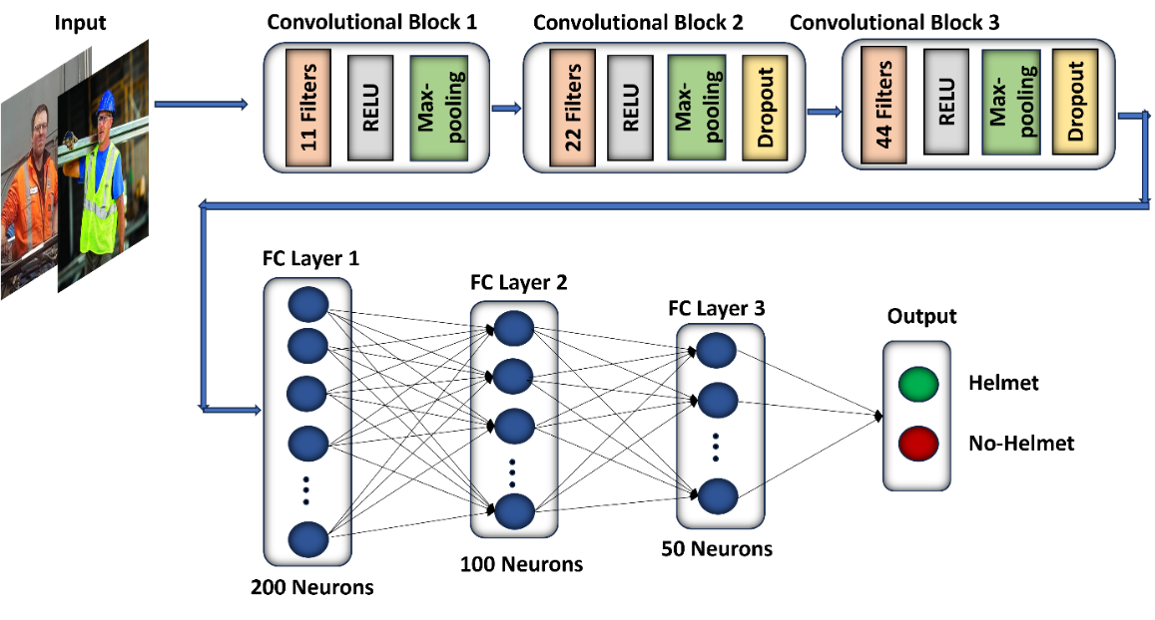}
    \caption{Schematic Representation of the Proposed CNN Architecture: This diagram illustrates the sequential layout and connectivity of convolutional, normalization, pooling, and fully connected layers designed for enhanced image classification performance.}
    \label{fig:figure8}
\end{figure}

Each convolutional layer includes a 3x3 convolution, optional batch normalization, ReLU activation, and max-pooling. Dropout regularization is strategically placed after the second and third blocks to prevent overfitting. The fully connected layers progressively decrease in size, featuring 200, 100, and 50 neurons, with the final layer culminating in a 2-neuron output for binary classification as can be seen in figure \ref{fig:figure8}.

Table \ref{tab:table4} provides a detailed breakdown of the internal architecture layout, describing each layer along with its corresponding output shape and the number of parameters, thereby elucidating the model's complexity and design:

\begin{table}[!htbp]
\centering
\caption{Detailed Specifications of Each Layer in the Proposed CNN Architecture.}
\begin{tabular}{|c|c|c|}
\hline
\textbf{Layers} & \textbf{Output Shape} & \textbf{Parameters} \\
\hline
Input & 3, 224x224 & - \\
\hline
Conv2d-1 & 222x222 & 308 \\
\hline
ReLU & 222x222 & - \\
\hline
MaxPool2d & 111x111 & - \\
\hline
Conv2d-2 & 109x109 & 2,178 \\
\hline
ReLU & 109x109 & - \\
\hline
MaxPool2d & 54x54 & - \\
\hline
Dropout & 54x54 & - \\
\hline
Conv2d-3 & 52x52 & 8,760 \\
\hline
ReLU & 52x52 & - \\
\hline
MaxPool2d & 26x26 & - \\
\hline
Dropout & 26x26 & - \\
\hline
FC1 & 200 Neurons & 5,950,000 \\
\hline
ReLU & 200 & - \\
\hline
Dropout & 200 & - \\
\hline
FC2 & 100 Neurons & 20,100 \\
\hline
ReLU & 100 & - \\
\hline
Dropout & 100 & - \\
\hline
FC3 & 50 Neurons & 5,050 \\
\hline
ReLU & 50 & - \\
\hline
Dropout & 50 & - \\
\hline
Output & 2 Neurons & 102 \\
\hline
\textbf{Total Parameters} & - & 5,995,698 \\
\hline
\end{tabular}
\label{tab:table4}
\end{table}

The streamlined design of this architecture, featuring a total of 5.99 million parameters, presents a significant efficiency advantage over more complex models like ResNet, which typically encompasses around 11.69 million parameters \cite{RN21}. This efficiency ensures more rapid model training and inference phases without sacrificing accuracy or performance.

\section{Experimental Results}
\subsection{Hyperparameter Tuning}
In our initial experiment, we meticulously tuned the hyperparameters to enhance the model's training dynamics and performance. The settings selected are summarized in Table \ref{tab:table5}.

\begin{table}[h]
\centering
\caption{Hyperparameters Setting for CNN Model Training}
\begin{tabular}{|c|c|}
\hline
\textbf{Hyperparameter} & \textbf{Value} \\
\hline
Batch Size & 32 \\
\hline
Learning Rate & 0.02 \\
\hline
Epochs & 60 \\
\hline
Optimizer & SGD-M \\
\hline
Momentum & 0.9 \\
\hline
\end{tabular}
\label{tab:table5}
\end{table}

\subsection{Initial Architecture Performance with Original Dataset}
The initial model architecture was assessed using the original, un-augmented dataset. This dataset, comprising 500 images split evenly across the two classes, presented a substantial challenge in terms of model generalization. The performance metrics are summarized in Table \ref{tab:table6}. During the training phase, the model achieved a peak training accuracy of 100\% but demonstrated a validation accuracy that fluctuated significantly, peaking at around 65\% in the early epochs and then stabilizing closer to 56.6\% by the 60th epoch. This variation and the notable gap between training and validation accuracy illustrate the initial model's struggles with overfitting and its inability to generalize effectively across the dataset.

\begin{table}[h]
\centering
\caption{Performance Metrics of the Initial CNN Model on the Original Dataset}
\begin{tabular}{|c|c|}
\hline
\textbf{Metric} & \textbf{Performance} \\
\hline
Precision & 52.0 \\
\hline
Recall & 57.0 \\
\hline
F1-Score & 55.0 \\
\hline
Accuracy & 65.0 \\
\hline
\end{tabular}
\label{tab:table6}
\end{table}

\begin{figure}[h]
    \centering
    \includegraphics[width=0.4\linewidth]{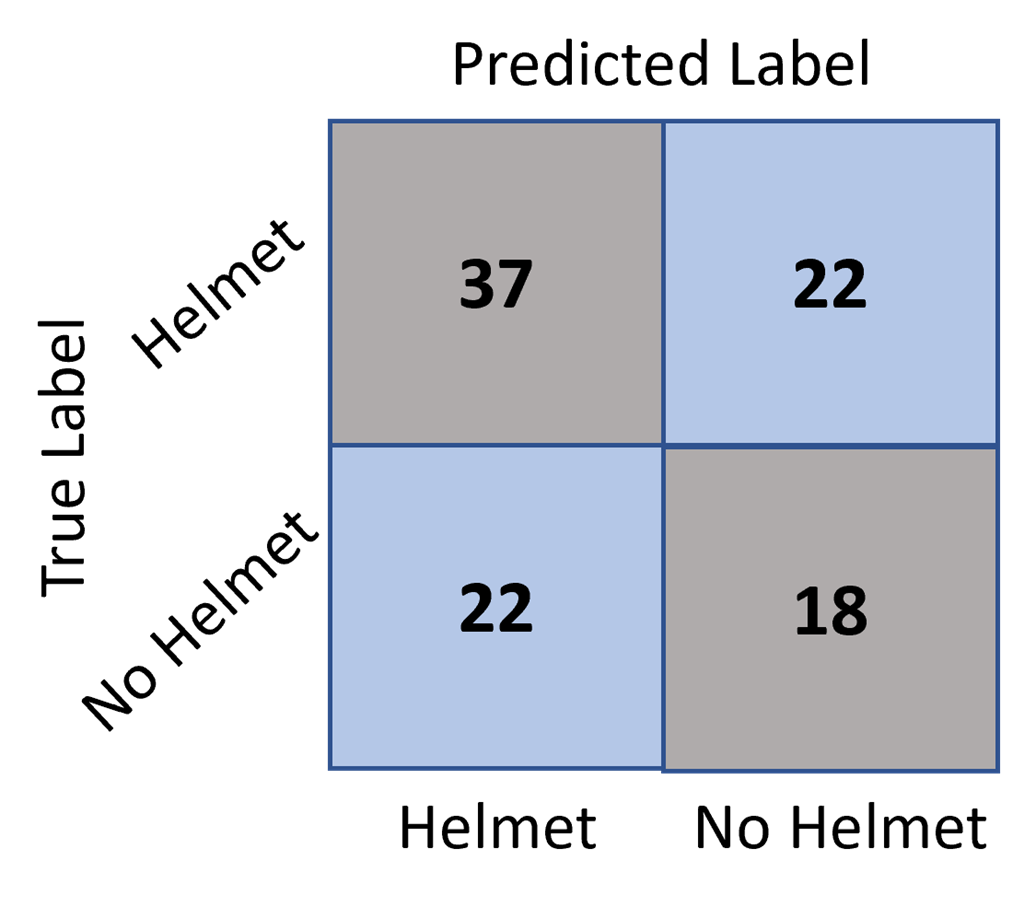}
    \caption{Detailed Confusion Matrix for the Initial Model Evaluated with the Original Dataset}
    \label{fig:figure10}
\end{figure}

The performance shortcomings of the initial architecture are further evidenced by the class-based confusion matrix (Figure \ref{fig:figure10}), which illustrates the misclassification rates within the dataset. Notably, 22 helmet-wearing samples were incorrectly labeled as 'no helmet', and similarly, 22 non-helmet samples were misclassified.

Given the limited dataset size and the initial model's struggles with generalization, enhancing the architecture's complexity was considered. However, to avoid excessive parameter increase and potential overfitting, we opted instead to explore data augmentation techniques as a strategy to improve the model's ability to generalize across varied scenarios.

\subsection{Initial Architecture Performance with Augmented Dataset}
Upon testing with the augmented dataset, the initial architecture demonstrated improved performance metrics, achieving a precision of 79\%, recall of 83\%, and both F1-score and accuracy of 81\% as detailed in Table \ref{tab:table7}. Despite these promising results, it should be noted that the same model only achieved 65\% accuracy with the original dataset, highlighting its susceptibility to overfitting—an issue quantified at 19\%. The validation accuracy of the model with the augmented data started at a high level and quickly stabilized, maintaining an average accuracy of approximately 81.8\% throughout the training process. This steady performance contrasts with the more variable accuracy observed with the original dataset, underscoring the benefits of data augmentation in enhancing model stability and generalization.

\begin{table}[h]
\centering
\caption{Evaluation Metrics for the Initial CNN Model Using the Augmented Dataset}
\begin{tabular}{|c|c|}
\hline
\textbf{Metric} & \textbf{Performance} \\
\hline
Precision & 79.0 \\
\hline
Recall & 83.0 \\
\hline
F1-Score & 81.0 \\
\hline
Accuracy & 81.0 \\
\hline
\end{tabular}
\label{tab:table7}
\end{table}

The observed performance prompted a transition to a refined architecture aimed at further enhancing model robustness and reducing overfitting.

\subsection{Modified Architecture Performance with Augmented Dataset}
The revised model structure, featuring two convolutional blocks and two fully connected layers, exhibited a slightly different performance profile. The results are summarized in Table \ref{tab:table8}, with the model achieving an F1-score and accuracy of 79\% and 80\%, respectively. The model's training accuracy quickly reached 100\% and remained consistent throughout the training process, while the validation accuracy demonstrated a rapid increase to around 80.76\% in the early epochs before stabilizing. This pattern suggests a slight increase in overfitting by 1\% over the initial model, as the high training accuracy was not fully mirrored in the validation performance.

\begin{table}[h]
\centering
\caption{Evaluation Metrics for the Modified CNN Model on Augmented Data}
\begin{tabular}{|c|c|}
\hline
\textbf{Metric} & \textbf{Performance} \\
\hline
Precision & 79.0 \\
\hline
Recall & 80.0 \\
\hline
F1-Score & 79.0 \\
\hline
Accuracy & 80.0 \\
\hline
\end{tabular}
\label{tab:table8}
\end{table}

Given the challenges associated with the increased overfitting, further exploration into advanced regularization techniques is warranted. Strategies such as batch normalization and dropout are being considered to mitigate the model's complexity and enhance its generalization capabilities. These techniques aim to reduce dependency on particular training examples and improve the model's predictive accuracy on unseen data.

\subsection{Batch Normalization}
The implementation of batch normalization in the modified model significantly enhanced its performance. This approach resulted in notable improvements across all metrics as summarized in Table \ref{tab:table9}. The model achieved a precision of 82\%, recall of 88\%, F1-score of 85\%, and an overall accuracy of 85\%. These results underscore the model's ability to deliver accurate and reliable predictions. The training accuracy of the model swiftly reached and maintained 100\%, while the validation accuracy, after an initial steep rise, stabilized at approximately 85.27\% throughout the majority of the epochs. This demonstrates a more consistent model behavior compared to previous iterations, with a reduced degree of overfitting now marked at only 15\%.

\begin{table}[h]
\centering
\caption{Performance Metrics of the Model Using Batch Normalization}
\begin{tabular}{|c|c|}
\hline
\textbf{Metric} & \textbf{Performance} \\
\hline
Precision & 82.0 \\
\hline
Recall & 88.0 \\
\hline
F1-Score & 85.0 \\
\hline
Accuracy & 85.0 \\
\hline
\end{tabular}
\label{tab:table9}
\end{table}

\subsection{Dropouts}
To further refine the model's performance, different dropout rates were tested, ranging from 10\% to 50\%. Table \ref{tab:table10} compares these configurations, showing minor fluctuations in performance metrics, with dropout proving to moderately influence model behavior. The precision varied between 81\% and 83\%, and recall between 85\% and 89\%. The F1-score remained consistently around 84\%, while training and validation accuracy exhibited variations correlated with the dropout rates.

\begin{table}[!htbp]
\centering
\caption{Comparative Analysis of Model Performance at Various Dropout Rates}
\begin{tabular}{|c|c|c|c|c|c|c|}
\hline
\textbf{Dropout Rate} & \textbf{Precision} & \textbf{Recall} & \textbf{F1-Score} & \textbf{Training Accuracy} & \textbf{Validation Accuracy} & \textbf{Overfitting} \\
\hline
10 & 81.0 & 89.0 & 84.0 & 100 & 85.0 & 15 \\
\hline
15 & 83.0 & 85.0 & 84.0 & 99.0 & 85.0 & 14 \\
\hline
25 & 82.0 & 85.0 & 84.0 & 99.0 & 84.0 & 13 \\
\hline
50 & 82.0 & 85.0 & 83.0 & 95.0 & 83.0 & 12 \\
\hline
\end{tabular}
\label{tab:table10}
\end{table}

\begin{figure}[h]
    \centering
    \includegraphics[width=0.75\linewidth]{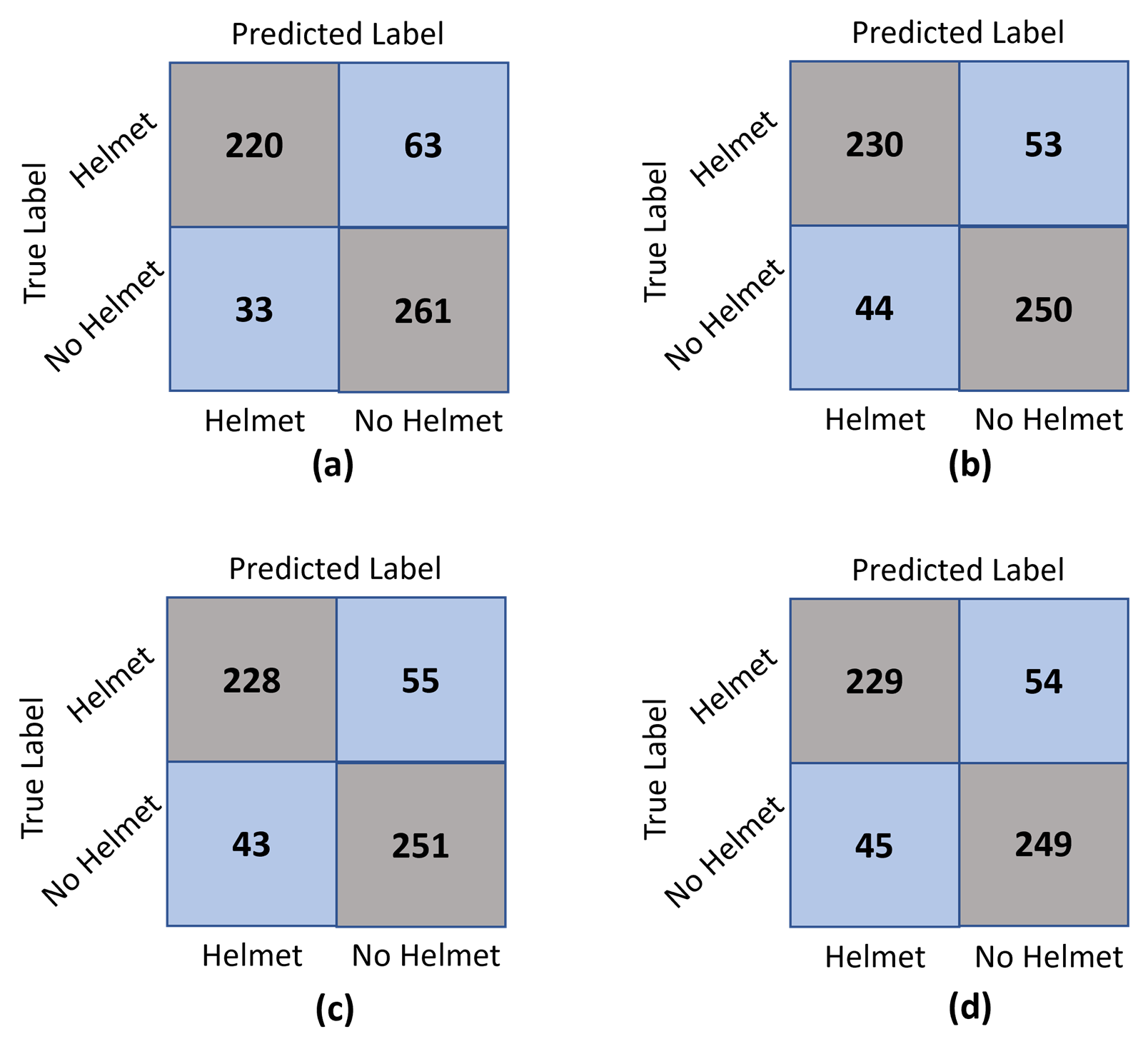}
    \caption{Visual Representation of Confusion Matrices for the Model at Various Dropout Rates: (a) 10\%, (b) 15\%, (c) 25\%, (d) 50\%.}
    \label{fig:figure14}
\end{figure}

These experimental results provide a detailed insight into how dropout rates affect the model's ability to generalize and reduce over-reliance on specific training samples, thus enhancing the robustness of predictions across unseen data.

\subsection{Batch Normalization and Dropout}
The integration of batch normalization alongside varying dropout rates was systematically evaluated to assess performance enhancements and overfitting characteristics. Table \ref{tab:table11} details the results, with the best performance observed at a 10\% dropout rate, which achieved a precision of 81\%, a recall of 90\%, and an F1-score of 85\%. This configuration also recorded the lowest degree of overfitting at 15\%, with the highest validation accuracy of 85\%. Other dropout rates—15\%, 25\%, and 50\%—yielded slightly lower metrics, each with a consistent overfitting degree of 16\% and similar validation accuracy.

\begin{table}[h]
\centering
\caption{Impact of Combining Batch Normalization with Different Dropout Rates on Model Performance}
\begin{tabular}{|c|c|c|c|c|c|c|}
\hline
\textbf{Batch Norm \& Dropout} & \textbf{Precision} & \textbf{Recall} & \textbf{F1-Score} & \textbf{Training Accuracy} & \textbf{Validation Accuracy} & \textbf{Overfitting} \\
\hline
 10\% & 81 & 90 & 85 & 100 & 85 & 15 \\
\hline
15\% & 82 & 86 & 84 & 100 & 84 & 16 \\
\hline
25\% & 82 & 85 & 84 & 100 & 84 & 16 \\
\hline
50\% & 82 & 85 & 83 & 99 & 83 & 16 \\
\hline
\end{tabular}
\label{tab:table11}
\end{table}

\begin{figure}[h]
    \centering
    \includegraphics[width=0.65\linewidth]{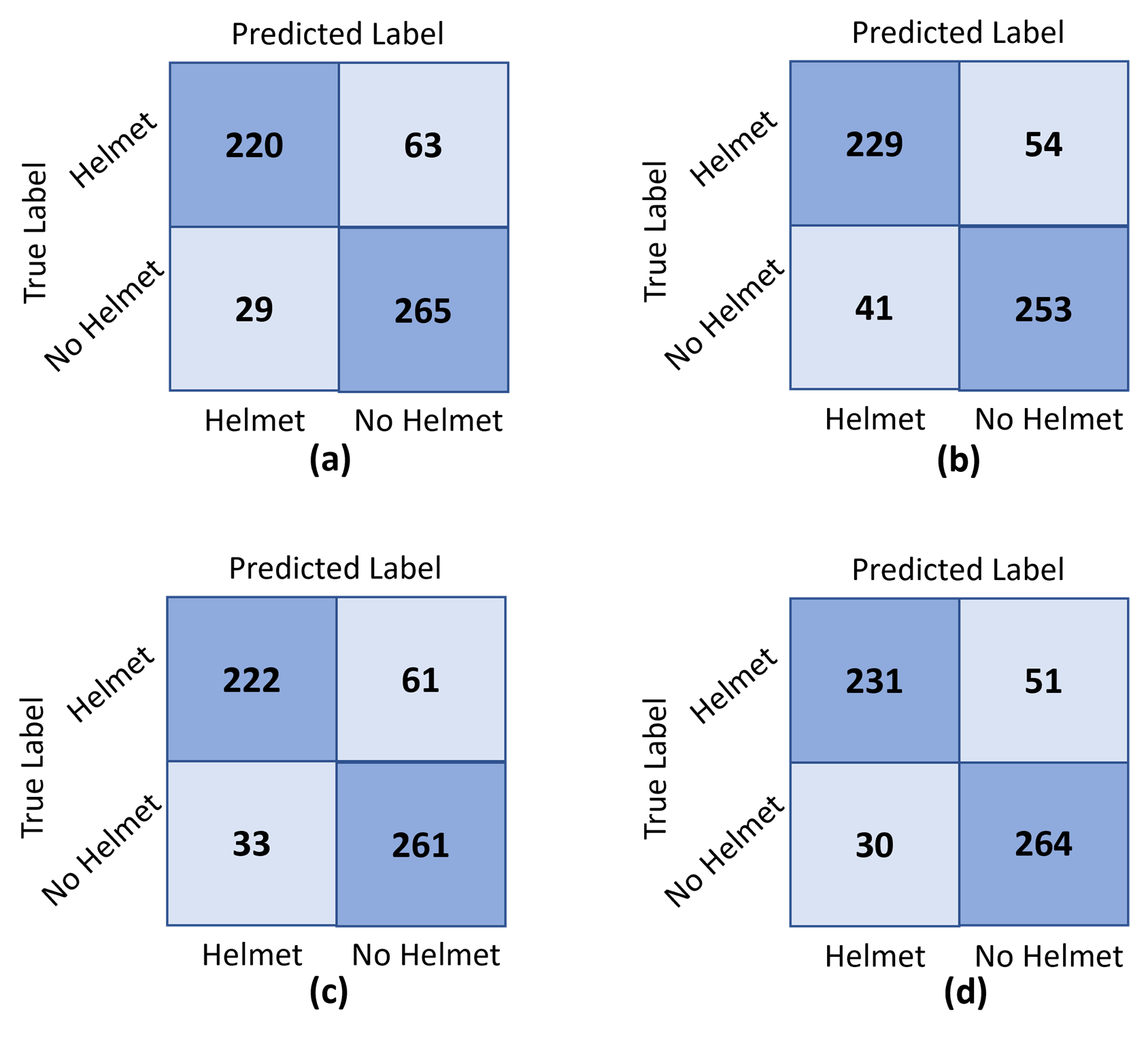}
    \caption{Confusion Matrices Illustrating the Impact of Batch Normalization Combined with Various Dropout Rates: (a) 10\%, (b) 15\%, (c) 25\%, (d) 50\%.}
    \label{fig:figure15}
\end{figure}

Figure \ref{fig:figure15} illustrates the confusion matrices corresponding to each dropout setting, providing a detailed view of classification performance. Specifically, the model with a 10\% dropout rate misclassified fewer images, while the model with a 50\% dropout rate showed the lowest misclassification rates for the 'helmet' category.

Despite the favorable outcomes with batch normalization and various dropout rates, it is important to recognize the persistent issue of overfitting. While the performance metrics were generally satisfactory, the notable degree of overfitting suggests that the model may be overly specialized to the training data and might not generalize well to new, unseen data. To mitigate this and enhance overall model performance, further explorations into architectural modifications and additional regularization techniques will be necessary.

\section{Discussion}
This study methodically evaluated the performance of two CNN architectures developed to classify helmet images within construction settings. The first model tested, an elementary architecture comprising one convolutional block and one fully connected layer, exhibited modest capabilities. On the original dataset, this initial model achieved a 55\% F1-score and a 65\% accuracy rate, underscoring substantial issues with overfitting, evident from a high overfitting rate of 35\%.

To rectify the observed shortcomings of the initial model, the architecture was enhanced by introducing an additional convolutional block and a second fully connected layer. This modified model was then augmented with batch normalization and varying dropout rates, aimed at boosting its effectiveness and minimizing overfitting. The performance metrics for the initial and modified models, detailed in Table \ref{tab:table16}, reveal the improvements achieved across various experimental setups, particularly when augmented data was employed.

\begin{table}[h]
\centering
\caption{Comparative Performance Metrics of Initial and Modified Models Under Various Experimental Conditions}
\begin{tabular}{|c|c|c|c|c|c|c|}
\hline
\textbf{Models} & \textbf{Experiments} & \textbf{Precision} & \textbf{Recall} & \textbf{F1-Score} & \textbf{Accuracy} & \textbf{Degree of Overfitting} \\
\hline
\multirow{2}{*}{Initial Model} & Original Dataset & 52 & 57 & 55 & 65 & 35 \\
                               & Augmented Dataset & 79 & 83 & 81 & 81 & 19 \\
\hline
\multirow{4}{*}{Modified Model} & Batch-Norm \& Dropout 10\% & 81 & 90 & 85 & 85 & 15 \\
                                & Batch-Norm \& Dropout 15\% & 82 & 86 & 84 & 84 & 16 \\
                                & Batch-Norm \& Dropout 25\% & 82 & 85 & 84 & 84 & 16 \\
                                & Batch-Norm \& Dropout 50\% & 82 & 85 & 83 & 83 & 16 \\
\hline
\end{tabular}
\label{tab:table16}
\end{table}

The results indicate that the modified model, particularly with the implementation of batch normalization and dropout at 10\% and 15\%, showed notable improvements. The inclusion of additional convolutional and fully connected layers, alongside these regularization techniques, significantly enhanced the model's ability to generalize, as demonstrated by its performance on the augmented dataset. These modifications not only improved the precision and recall but also effectively reduced the degree of overfitting to 15\% and 16\%, respectively.

This exploration confirms the benefits of incremental architectural enhancements and sophisticated regularization strategies in refining the predictive accuracy and generalizability of CNNs for helmet detection. Such advancements are crucial for deploying robust models in real-world construction environments where variability in conditions can challenge the efficacy of simpler models.

\section{Conclusion}
This paper has methodically explored the initial development and incremental improvement of convolutional neural networks (CNNs) for helmet detection in construction environments. We began with a basic CNN model consisting of one convolutional block and one fully connected layer. Subsequent enhancements, including the addition of another convolutional block and a second fully connected layer, along with the integration of batch normalization and dropout techniques, have significantly bolstered the model's performance.

Through these modifications, the models showed improved capability to classify helmet images accurately. The modified architecture, incorporating advanced regularization techniques, achieved notable enhancements in precision, recall, and F1-score, effectively addressing the initial model's limitations of underfitting and excessive overfitting \cite{hussain2022statistical}.

\section{Future Work}
While this study has laid a solid foundation for helmet detection using CNNs, further advancements are necessary to optimize and expand the current model's capabilities:
\begin{itemize}
    \item \textbf{Real-Time Deployment:} Future work should aim at optimizing the model for real-time application, ensuring it can operate effectively in dynamic construction site along with other environments such as renewable energy \cite{hussain2019deployment}.
    \item \textbf{Integration with IoT Devices:} Integrating the CNN model with IoT devices could lead to the development of comprehensive safety monitoring systems, enhancing proactive safety measures on construction sites \cite{aydin2023domain}.
    \item \textbf{Expansion to Other PPE Detection:} Extending the model's capabilities to detect other personal protective equipment, such as safety vests and goggles, could provide a more holistic approach to site safety.
    \item \textbf{Robustness to Environmental Variabilities:} Enhancing the model to maintain high accuracy across diverse environmental conditions, including different lighting and weather conditions, remains a priority.
    \item \textbf{Data Augmentation:} Enriching the training dataset with a wider variety of images, including those from different geographic locations and varied construction settings, would help improve the model's robustness and generalizability.
    \item \textbf{Exploration of Advanced Architectures:} Investigating more sophisticated neural network architectures and further refining dropout and normalization strategies could enhance the model's efficiency and effectiveness. Explorations in lightweight CNN architectures for digital applications and child emotion recognition present exciting avenues \cite{hussain2023child}.
\end{itemize}

By addressing these areas, future research can significantly advance the state of automated safety monitoring in construction and similar industrial sectors, paving the way for safer working conditions through enhanced technological integration.

\bibliographystyle{unsrt}  
\bibliography{references}

\end{document}